**Efficient Transformer-Based Localized Patch Sampling for Choroid Plexus Segmentation in Multiple Sclerosis**

Po-Jui Lu[1,2,3], Alessandro Cagol[1,2,3,4], Mario Ocampo-Pineda[1,2,3], Federico Spagnolo[1,2,3], Marina Mastantuono, Msc[1,5], Andreea-Alexandra Aldea[1,2,3], Jannis Müller, MSc[1,2,3], Özgür Yaldizli[1,2,3], Matthias Weigel[1,2,3,6], Lester Melie-Garcia[1,2,3], Roberta Magliozzi[5,7], Maria Pia Sormani[4,8], Ludwig Kappos[1,2,3], Jens Kuhle[2,3], Cristina Granziera[1,2,3]

1. *Translational Imaging in Neurology (ThINk) Basel, Department of Biomedical Engineering, Faculty of Medicine, University Hospital Basel and University of Basel, Basel, Switzerland*
2. *Department of Neurology, University Hospital Basel, Basel, Switzerland*
3. *Research Center for Clinical Neuroimmunology and Neuroscience Basel (RC2NB), University Hospital Basel and University of Basel, Basel, Switzerland*
4. *Dipartimento di Scienze della Salute, Università degli Studi di Genova, Italy*
5. *Neurology Section of Department of Neurological and Movement Sciences, University of Verona, Italy*
6. *Division of Radiological Physics, Department of Radiology, University Hospital Basel, Basel, Switzerland*
7. *Department of Brain Sciences, Faculty of Medicine, Imperial College London, United Kingdom*
8. *IRCCS Ospedale Policlinico San Martino, Genova, Italy*

**Abstract**

**Background**

The lateral ventricle choroid plexus (LVCP) is gaining recognition as a key imaging biomarker for multiple sclerosis (MS) related to physical disability and neuroinflammation. Yet, manual segmentation of the LVCP is highly tedious, restricting its use in broad clinical trials and longitudinal assessments.

**Purpose**

This research aims to develop and test a SwinUNETR-driven pipeline that leverages targeted intra- and peri-ventricular small patch sampling to automatically segment the LVCP in MS patients from both standalone and multi-modal MRI inputs.

**Methods**

We retrospectively assessed 3T MRI scans across three sets of data stemming from two separate MS-dominant cohorts (Dataset 1: n=177; Dataset 2: n=177; expanded test set: n=388). Our method employed a SwinUNETR architecture trained on 32×32×32 voxel patches, benchmarking it against the 3D UXNET model. The primary metric for evaluation was the Dice Similarity Coefficient (DSC), supplemented by computational demand (GFLOPs) and the 95th percentile Hausdorff Distance (HD95).

**Results**

On the extended test set of 388 patients, the SwinUNETR model secured a mean DSC of 0.868 (95% CI: 0.863-0.872) with MPRAGE and FLAIR combined, showing a statistically significant gain over UXNET (DSC: 0.858 [95% CI: 0.853-0.862], p<0.0001). When restricted to standalone FLAIR inputs, the transformer-based approach sustained a high DSC of 0.863 (95% CI: 0.858-0.867), while the spatial localization of UXNET worsened considerably (HD95: 1.86 vs. 3.00 mm). Importantly, the proposed framework lowered computational load by 99% (91.8 vs. 22,080 GFLOPs) and operated with significantly fewer parameters (4.08M vs. 53.02M).

**Conclusion**

By integrating localized patch sampling with a SwinUNETR architecture, this methodology offers an accurate, robust, and statistically superior alternative to current leading models for LVCP segmentation. Its vast reduction in computational cost makes it ideal for widespread implementation in clinical and research environments.

## Introduction

The choroid plexus (CP), a highly vascularized network situated in the ventricular system, plays a crucial role in cerebrospinal fluid generation and acts as a gateway for immune cells entering the central nervous system (1,2). Emerging MRI research highlights noticeable CP expansion in individuals with multiple sclerosis (MS), as well as Parkinson's and Alzheimer's diseases (2–4). Specifically in MS, histopathological studies reveal inflammatory cells within the CP (5) and the volume of the lateral ventricle CP (LVCP) is linked to diffuse neurodegeneration and both physical and cognitive impairment (6–10), making it a promising imaging biomarker. Nevertheless, CP structure naturally shifts with age, beginning around the third decade of life (11). Because manual segmentation of the LVCP is highly time-consuming, evaluating CP metrics in extensive prospective trials and longitudinal studies remains a major challenge.

Various deep learning frameworks have been developed for LVCP segmentation. ASCHOPLEX (12) is an U-Net-based (13) ensemble, achieving a mean test Dice Similarity Coefficient (DSC) of 0.80 on Magnetization Prepared RApid Gradient Echo (MPRAGE) images from healthy and MS cohorts, much better than FreeSurfer (14). Another UNETR-based approach, utilizing intra- and peri-ventricular patch sampling, outperformed ASCHOPLEX (test DSC: 0.77 vs 0.70) in MS cohorts and demonstrated improved performance with multi-contrast integration (e.g., MPRAGE/FLAIR) (DSC: 0.84) compared with single-modality MPRAGE (15). While these models utilize standard transformer backbones, SwinUNETR has demonstrated superior performance in brain tumor segmentation by leveraging multi-scale feature extraction (16). Recently, the convolution-based UXNET (17) reported a test DSC of 0.875 on two-timepoint MPRAGE images from people with MS (18); however, its validation was limited to a relatively young cohort (31.1±9.4 years). Additionally, UXNET's reliance on large patch sizes imposes substantial GPU memory overhead, potentially limiting its clinical deployment. Beyond MS, U-Net-based ChpSeg (19), was trained on single-modality T1-weighted, T2-weighted, or FLAIR images from Parkinson's and Alzheimer's cohorts (DSC: 0.74).

To address these challenges, this work introduces an automated LVCP segmentation pipeline that combines a SwinUNETR backbone with a targeted intra- and peri-ventricular small patch sampling strategy. We trained these models from scratch leveraging two separate MS-predominant cohorts (which included related demyelinating and neuroinflammatory conditions): one utilizing FLAIR and MPRAGE, and the second utilizing FLAIR and MP2RAGE. We assessed the framework's efficacy across both multi-modal and single-modality inputs, benchmarking the results against the UXNET architecture. Finally, the generalizability and robustness of the methodology were comprehensively validated using a large, independent extended test set.

## Materials and Methods

### Participants and Imaging Protocol

The study retrospectively included 1-mm isotropic 3T 3D MRI from two prospective MS-predominant cohort studies (NCT02433028, hereafter Cohort 1; NCT05177523, hereafter Cohort 2) at the University Hospital of Basel. Both studies were approved by the local ethics review committee (Ethics Committee Northwest and Central Switzerland, BASEC-ID: 2018-01174 and 2023-02367), and all participants provided written consent prior to the studies. Dataset sizes derived from the two cohorts were matched.

Dataset 1 included 177 participants with FLAIR and MPRAGE images randomly sourced from Cohort 1. The inclusion criteria were: (1) age ≥ 18 years; (2) ability to provide informed consent; (3) MS diagnosis according to the 2017 McDonald Criteria, with no psychiatric or neurologic comorbidities, contraindications for MRI, and pregnancy; (4) MRI acquired at the University Hospital Basel; (5) not being in Cohort 2. The included participants were 150 relapsing remitting MS, 14 secondary progressive MS, 4 primary progressive MS, 7 clinically isolated syndrome , 1 radiologically isolated syndrome, and 1 neuromyelitis optica spectrum disorder (age: 48.8±11.8; 124 females; median[IQR] EDSS: 2.5[1.5-3.5]).

Dataset 2 consisted of 177 participants from Cohort 2 (age: 46.6±14.0; 106 females; median[IQR] EDSS:3.0[2.0;4.5]; 98 relapsing remitting MS, 58 secondary progressive MS, and 21 primary progressive MS) undergoing an MRI protocol including FLAIR and MP2RAGE. Participants met the same inclusion criteria (1-4) as Dataset 1.

To evaluate model robustness on a larger scale, an independent subset of 388 participants from Cohort 1 was utilized as an extended test dataset, following the same inclusion criteria of Dataset 1 (age: 44.1±12.7; 252 females; median[IQR] EDSS: 2.0[1.5;3.5]; 338 relapsing remitting MS, 18 secondary progressive MS, 16 primary progressive MS, 6 clinically isolated syndrome, 1 myelin oligodendrocyte glycoprotein antibody-associated disease, 4 neuromyelitis optica spectrum disorder, and 5 radiologically isolated syndrome).

Detailed MRI acquisition parameters are given in the Supplementary Material.

**Image Preprocessing**

The Cohort 1 image processing pipeline included skull-stripping through HD-BET (20), rigid co-registration of FLAIR and MPRAGE images using ANTs (21), and segmentation using SAMSEG (22). Skull-stripped, registered FLAIR and MPRAGE images, as well as ventricle mask, were obtained.

For Cohort 2, HD-BET skull-stripping, rigid co-registration of FLAIR and MP2RAGE images using FSL (23) and SAMSEG brain segmentation were used.

**Dataset Preparation**

Dataset 1 (n=177) was randomly divided into a held-out test dataset (n=36) and a development dataset (n=141) for five-fold cross-validation. Dataset 2 followed the identical protocol.

**Choroid Plexus Manual Segmentation**

For Datasets 1 and 2, initial CP masks were created using SAMSEG. Two raters manually corrected 10 CP masks on FLAIR/MPRAGE images using ITK-Snap (24), achieving high inter-rater agreement (DSC: 0.877), and corrected the rest of masks accordingly and independently.

To facilitate manual correction for the extended test dataset, a UNETR model was trained on Dataset 1 following the approach described in (15). This model achieved a DSC>0.8. The resulting masks were subsequently manually corrected by the raters following the same criteria applied to Dataset 1.

**Model Training**

For the SwinUNETR, the number of attention heads was set to (3,6,12,24), with an initial embedding dimension of 12. Instance normalization was omitted from the final encoder stage to accommodate a minimal input patch size of 32×32×32 voxels. At this stage of the original design, the dimension of the hidden feature representation was 1x1x1, resulting in a zero-variance denominator for instance normalization.

Voxel intensities were normalized within the brain using z-scores. Intra- and peri-ventricular patches were sampled within the bounding box of the ventricle zero-padded by 8 voxels in all directions. Stochastic data augmentation included cropping, flipping, intensity shift of 0.1, and affine transformations with up to 5% scaling and up to 12° rotation. The loss function was a combination of the Focal loss (25) and the Dice loss. The model was optimized by AdamW (26) with DSC as the evaluation metric.

The code was implemented with Python (3.10; pytorch v2.0) and R (4.5.1).

**Statistical Analysis**

Segmentation performance was evaluated using the DSC as the primary outcome measure. To provide a comprehensive morphological assessment, the 95th percentile Hausdorff Distance (HD95), precision, recall, and volume similarity (27) were calculated as secondary measures. Comparative analyses evaluated the impact of single-modality versus multi-modal MRI input (MPRAGE, FLAIR, and MPRAGE+FLAIR), patch size selection, and bias-field correction. In addition, the impact of different ventricle masks on the inference was also examined (SAMSEG vs FastSurfer (28)).

UXNET was also trained across identical contrast combinations of full-brain images. Notably, the minimal patch size acceptable for UXNET was 96x96x96 voxels. Computational efficiency was quantified through total giga floating-point operation counts (GFLOPs). Statistical significance for performance differences in the primary measure across the models and MR contrasts, was assessed using a two-tailed Wilcoxon signed-rank test. The 95% confidence intervals (CIs) for both absolute measures and paired performance differences were computed using 10000-iteration bootstrapping.

**Segmentation Performance on Dataset 1**

The initial evaluation of model architectures and imaging contrasts was performed using cross-validation on the development dataset of Dataset 1 (Table 1). SwinUNETR utilizing multi-modal MPRAGE+FLAIR inputs achieved the highest mean DSC of 0.866, demonstrating superior performance compared with the MPRAGE-only input (DSC 0.790). A similar trend was observed for the UXNET architecture, where the MPRAGE-only input led to a 10% reduction in DSC to 0.767. Accordingly, the MPRAGE-only input was excluded from further evaluation on the test datasets to focus on the highest-performing modalities.

**Independent and Extended Test Set Validation**

SwinUNETR (MPRAGE+FLAIR) maintained the highest mean DSC of 0.877 (95% CI: 0.863- 0.888) on the held-out test set and 0.868 (95% CI: 0.863-0.872) on the extended test dataset (Table 1). Based on the primary outcome measure, SwinUNETR significantly outperformed UXNET across the extended test dataset when using MPRAGE+FLAIR (mean paired difference: 0.010 [95% CI: 0.007-0.012]; Wilcoxon signed-rank test, $n=388$, $V=64071$, $p<0.0001$), and when using FLAIR only (mean paired difference: 0.005 [95% CI: 0.003-0.007]; $n=388$, $V=53575$, $p<0.0001$). Figures 1 and 2 show LVCP segmentation masks created by the SwinUNETR and UXNET using MPRAGE+FLAIR on two people with MS with different sizes of lateral ventricles. While SwinUNETR and UXNET performed comparably in HD95 and volume similarity using MPRAGE+FLAIR, SwinUNETR exhibited a better precision-recall balance (0.864 and 0.879, respectively) than UXNET (0.848 and 0.876, respectively). Furthermore, the HD95 of the UXNET deteriorated from 1.77±1.73 mm to 3.0±6.93 mm when only FLAIR was used as the input.

| | | **Five-fold Cross-Validation Dataset (n=141)** | **Test Dataset (n=36)** | **Extended Test Dataset (n=388)** | | | | |
|---|---|---|---|---|---|---|---|---|
| **Neural Network** | **Contrasts** | **DSC** | **DSC [95% CI]** | **DSC [95% CI]** | **HD95 [95% CI]** | **Precision [95% CI]** | **Recall [95% CI]** | **Volume Similarity [95% CI]** |
| **Swin UNETR** | MPRAGE +FLAIR | 0.866±0.014† | 0.877±0.038 [0.863, 0.888] | 0.868±0.048 [0.863, 0.872] | 1.84±2.58 [1.65, 2.20] | 0.863±0.072 [0.856, 0.870] | 0.879±0.065 [0.872, 0.885] | 0.951±0.037 [0.947, 0.954] |
| | FLAIR | 0.866±0.016† | 0.878±0.038 [0.864, 0.889] | 0.863±0.049 [0.858, 0.867] | 1.86±2.32 [1.67, 2.15] | 0.869±0.071 [0.862, 0.876] | 0.863±0.067 [0.856, 0.869] | 0.950±0.037 [0.946, 0.954] |
| | MPRAGE | 0.790±0.014† | - | - | - | - | - | - |
| **UXNET** | MPRAGE +FLAIR | 0.861±0.010† | 0.870±0.042 [0.854, 0.881] | 0.858±0.046 [0.853, 0.862] | 1.77±1.73 [1.62, 1.99] | 0.848±0.078 [0.840, 0.855] | 0.876±0.057 [0.870, 0.882] | 0.948±0.043 [0.943, 0.952] |
| | FLAIR | 0.861±0.016† | 0.870±0.042 [0.855, 0.882] | 0.852±0.055 [0.846, 0.858] | 3.00±6.97 [2.43, 3.87] | 0.828±0.076 [0.820, 0.835] | 0.885±0.069 [0.877, 0.892] | 0.946±0.044 [0.941, 0.950] |
| | MPRAGE | 0.767±0.017† | - | - | - | - | - | - |

Table 1. Performance measures. The mean and standard deviation of performance measures are reported. The standard deviation across validation folds is marked with †. DSC stands for Dice Similarity Coefficient, where larger values indicate better performance. HD95 is 95th percentile Hausdorff distance, where smaller values represent better performance. The dash sign indicates the model was excluded from testing due to suboptimal cross-validation performance.

**Comparative Analysis**

The results of the analyses investigating the impact of input patch size, bias-field correction, and ventricle mask source are summarized in Table 2.

Initial optimization focused on input patch size using cross-validation on the development dataset. The 32×32×32 voxel configuration achieved a mean DSC of 0.866, comparable to the larger 64x64x64 patch size, but superior to the 16×16×16 patch size. Consequently, the 32×32×32 configuration was selected as the optimal baseline for subsequent experiments. With this optimized patch size, the impact of bias-field correction was evaluated. The mean DSC remained stable between uncorrected (0.866±0.014) and bias-field corrected inputs (0.864±0.013).

To assess the compatibility of the trained SwinUNETR with ventricle masks derived from different sources, the best-performing model from the five-fold cross-validation (trained with SAMSEG-derived masks) was evaluated on the held-out test dataset using ventricle masks from SAMSEG and FastSurfer. The DSCs for using FastSurfer- and SAMSEG-derived ventricle masks were comparable.

The computational efficiency and architectural complexity of the proposed SwinUNETR framework were compared against the UXNET model using combined MPRAGE and FLAIR inputs (Table 3). The SwinUNETR demonstrated a substantially lower computational cost measured in GFLOPs and a marked reduction in the total number of trainable parameters.

| | Five-fold Cross-Validation Development Dataset (n=141) | | | | Test Dataset (n=36) | |
|---|---|---|---|---|---|---|
| | **Input patch size** | **Mean Dice** | **Without Bias-field correction** | **With Bias field correction** | **SAMSEG ventricle mask** | **FastSurfer ventricle mask** |
| **SwinUNETR on MPRAGE+FLAIR** | 32*32*32 | 0.866±0.014† | 0.866±0.014† | 0.864±0.013† | 0.877±0.037 | 0.873±0.040 |
| | 16*16*16 | 0.836±0.015† | - | - | - | - |
| | 64*64*64 | 0.864±0.013† | - | - | - | - |

Table 2. Performance measures in the comparative analysis. The mean and standard deviation of performance measures are reported. The standard deviation across validation folds is marked with †. DSC stands for Dice Similarity Coefficient, where larger values indicate better performance. The dash sign indicates the model was excluded from further comparison due to suboptimal cross-validation performance.

| | Contrasts | Total GFLOPs | #Params |
|---|---|---|---|
| **SwinUNETR** | MPRAGE+FLAIR | 91.8 | 4.08M |
| **UXNET** | MPRAGE+FLAIR | 22080 | 53.02M |

Table 3. Efficiency comparison. Total GFLOPs stand for total giga floating point operation counts, where the smaller the value, the less the computation and the faster the inference. #Params shows the number of trainable parameters.

### Segmentation Performance on Dataset 2

Based on the optimizations established in the preceding subsections, this assessment focused exclusively on the dual-modality input configuration (MP2RAGE and FLAIR) (Table 4). The UXNET model achieved a marginal 0.2% of improvement in DSC in cross-validation and on the test dataset. The results across Dataset 2 showed the robust generalizability of both UXNET and SwinUNETR.

| | | **Five-fold Cross-Validation Dataset (n=141)** | **Test Dataset (n=36)** |
|---|---|---|---|
| **Neural Network** | **Contrasts** | **DSC** | **DSC** |
| **SwinUNETR** | MP2RAGE+FLAIR | 0.824±0.009† | 0.837±0.041 [0.823, 0.849] |
| **UXNET** | MP2RAGE+FLAIR | 0.826±0.007† | 0.839±0.036 [0.827, 0.850] |

Table 4. Performance measures of Dataset 2. The mean and standard deviation of performance measures are reported. The standard deviation across validation folds is marked with †. DSC stands for Dice Similarity Coefficient, where larger values indicate better performance.

## Discussion

This study demonstrates that a SwinUNETR-based framework with localized small-patch sampling provides a highly efficient and robust solution for automated LVCP segmentation in people with MS and related neuroinflammatory and demyelinating disorders. In addition, our findings indicate that multimodal MRI inputs are beneficial for achieving high, stable segmentation performance for both transformer-based and convolution-based neural networks.

Interestingly, SwinUNETR exhibited a minor performance decline when FLAIR was used as the sole input modality. This robustness supports its use in practical scenarios where only FLAIR is available. In contrast, UXNET's spatial localization accuracy, as measured by HD95, greatly deteriorated under this single-modality configuration. This suggests that the transformer's global self-attention mechanism combined with small patches may be better suited for capturing LVCP morphology.

Our results also demonstrate that specialized sampling strategies can help bridge the gap between model complexity and clinical feasibility. By utilizing intra- and peri-ventricular patch sampling, we focused the model's learning capacity on anatomically relevant regions. This allowed for the use of smaller 32x32x32 voxel patches, which, in combination with the SwinUNETR architecture, resulted in a framework with over 99% fewer GFLOPs than the convolution-based UXNET did. This reduction in computational overhead is essential for deploying such tools in standard clinical workstations, where GPU memory may be limited.

The minimal performance difference observed with bias-field correction for preprocessing further supports the robustness of the proposed framework. Moreover, the finding that SwinUNETR inference performance was independent of the ventricle mask source suggests no need for re-training when alternative ventricle segmentation methods are used, thereby enhancing the generalizability of the approach.

When comparing our outcomes to existing literature, SwinUNETR outperformed previously reported benchmarks, including ASCHOPLEX (DSC: 0.80) and UNETR-based approaches (DSC: 0.84). Although UXNET has recently reported a high DSC of 0.875, that study was limited to a young PwMS cohort. In contrast, our framework achieved comparable performance in an older MS population with a wider age range, which is critical given that CP morphology and the surrounding brain environment undergo significant changes across the adult lifespan. In our comparison, UXNET could achieve a DSC of 0.858 in the extended test dataset, while the proposed SwinUNETR framework achieved a DSC of 0.868 with a much lower computational load.

Validation on Dataset 2 with the MP2RAGE+FLAIR configuration revealed a moderate decrease in DSC for both SwinUNETR and UXNET (DSC ~0.82–0.83). This relative performance decrease may be attributable to inherent differences in image contrast and noise characteristics between the two T1-weighted sequences (MPRAGE and MP2RAGE). However, further studies with both sequences are required to confirm this hypothesis.

Some limitations warrant consideration. First, while we validated the model on two distinct cohorts acquired with different MRI systems, all data originated from a single clinical center. Future work should evaluate the model's performance on multicenter data to confirm robustness across hardware variations. Second, while the model inference is robust, generating the ventricle mask for sampling requires initial image processing. A low-resource object detection model, such as YOLO (29), could potentially provide a solution by rapidly identifying the ventricular ROI for patch extraction without full segmentation. Finally, the impact of motion artifacts on segmentation performance was not assessed in this study.

**Conclusion**

To summarize, this work presents and validates a highly efficient pipeline based on SwinUNETR for the automated segmentation of the LVCP. The results establish that combining T1-weighted and FLAIR MRI provides optimal accuracy, yet relying solely on FLAIR imaging is still a highly viable option. By drastically cutting down computational overhead while outperforming state-of-the-art techniques on a large-scale test set, the proposed method enables extensive volumetric analysis of the LVCP as a potential biomarker for neurodegeneration and neuroinflammation in clinical settings.

## References

1. Kolahi S, Zarei D, Issaiy M, Shakiba M, Azizi N, Firouznia K. Choroid plexus volume changes in multiple sclerosis: insights from a systematic review and meta-analysis of magnetic resonance imaging studies. Neuroradiology. 2024;c:1869–1886. doi: 10.1007/s00234-024-03439-3. Cited in PMID: 39105769.

2. Jankowska A, Chwojnicki K, Grzywińska M, Trzonkowski P, Szurowska E. Choroid Plexus Volume Change—A Candidate for a New Radiological Marker of MS Progression. Diagnostics. 2023;13. doi: 10.3390/diagnostics13162668.

3. Čarna M, Onyango IG, Katina S, Holub D, Novotny JS, Nezvedova M, Jha D, Nedelska Z, Lacovich V, Vyvere T Vande, et al. Pathogenesis of Alzheimer’s disease: Involvement of

the choroid plexus. Alzheimer's and Dementia. 2023;19:3537–3554. doi: 10.1002/alz.12970. Cited in PMID: 36825691.

4. Jeong SH, Park CJ, Jeong HJ, Sunwoo MK, Ahn SS, Lee SK, Lee PH, Kim YJ, Sohn YH, Chung SJ. Association of choroid plexus volume with motor symptoms and dopaminergic degeneration in Parkinson's disease. Journal of Neurology, Neurosurgery and Psychiatry. 2023;94:1047–1055. doi: 10.1136/jnnp-2023-331170. Cited in PMID: 37399288.

5. Magliozzi R, Hametner S, Mastantuono M, Mensi A, Karimian M, Griffiths L, Watkins LM, Poli A, Berti GM, Barusolo E, et al. Neuropathological and cerebrospinal fluid correlates of choroid plexus inflammation in progressive multiple sclerosis. Brain Pathology. 2024;1–20. doi: 10.1111/bpa.13322.

6. Preziosa P, Pagani E, Meani A, Storelli L, Margoni M, Yudin Y, Tedone N, Biondi D, Rubin M, Rocca MA, et al. Chronic Active Lesions and Larger Choroid Plexus Explain Cognition and Fatigue in Multiple Sclerosis. Neurol Neuroimmunol Neuroinflamm. 2024;11:e200205. doi: 10.1212/NXI.0000000000200205.

7. Barbuti E, Conti A, Treaba CA, Miscioscia A, Barletta VT, Herranz E, Sloane JA, Klawiter EC, Toschi N, Mainero C. Choroid Plexus Enlargement in Multiple Sclerosis Correlates with Cortical and Phase Rim Lesions on 7T MRI and Predicts Progression Independent of Relapse Activity. AJNR Am J Neuroradiol. 2026;47:557–565. doi: 10.3174/ajnr.A8983.

8. Ricigliano VAG, Morena E, Colombi A, Tonietto M, Hamzaoui M, Poirion E, Bottlaender M, Gervais P, Louapre C, Bodini B, et al. Choroid Plexus Enlargement in Inflammatory Multiple Sclerosis: 3.0-T MRI and Translocator Protein PET Evaluation. Radiology. 2021;301:166–177. doi: 10.1148/radiol.2021204426.

9. Müller J, Noteboom S, Granziera C, Schoonheim MM. Understanding the Role of the Choroid Plexus in Multiple Sclerosis as an MRI Biomarker of Disease Activity. Neurology. 2023;100:405–406. doi: 10.1212/WNL.0000000000206806. Cited in PMID: 36543568.

10. Müller J, Sinnecker T, Wendebourg MJ, Schläger R, Kuhle J, Schädelin S, Benkert P, Derfuss T, Cattin P, Jud C, et al. Choroid Plexus Volume in Multiple Sclerosis vs Neuromyelitis Optica Spectrum Disorder: A Retrospective, Cross-sectional Analysis. Neurol Neuroimmunol Neuroinflamm. 2022;9:e1147. doi: 10.1212/NXI.0000000000001147. Cited in PMID: 35217580.

11. Li J, Gao Y, Xu Y, Dai W, Hu Y, Feng X, Wu D, Zhao L. Morphological changes of the choroid plexus in the lateral ventricle across the lifespan: 5551 subjects from fetus to elderly. NeuroImage. 2025;318:121392. doi: 10.1016/j.neuroimage.2025.121392.

12. Visani V, Veronese M, Pizzini FB, Colombi A, Marjin C, Tamanti A, Schubert JJ, Althubaity N, Harrison NA, Bullmore ET, et al. ASCHOPLEX : a generalizable approach for the automatic segmentation of choroid plexus. 2024;182. doi: 10.1016/j.compbiomed.2024.109164.

13. Ronneberger O, Fischer P, Brox T. U-Net: Convolutional Networks for Biomedical Image Segmentation [Internet]. arXiv; 2015 [cited 2025 Nov 24]. Available from: https://arxiv.org/abs/1505.04597.

14. Fischl B. FreeSurfer. NeuroImage. 2012;62:774–781. doi: 10.1016/j.neuroimage.2012.01.021.

15. Lu P-J, Cagol A, Ocampo-Pineda M, Mastantuono M, Aldea A-A, Müller J, Yaldizli Ö, Weigel M, Melie-Garcia L, Magliozzi R, et al. Deep learning-based Segmentation of the Choroid Plexus in Multiple Sclerosis Using MPRAGE, FLAIR, and MP2RAGE. Mult Scler. 2025;31:1032. doi: 10.1177/13524585251358343.

16. Hatamizadeh A, Nath V, Tang Y, Yang D, Roth H, Xu D. Swin UNETR: Swin Transformers for Semantic Segmentation of Brain Tumors in MRI Images [Internet]. arXiv; 2022 [cited 2026 Feb 4]. Available from: http://arxiv.org/abs/2201.01266.

17. Lee HH, Bao S, Huo Y, Landman BA. 3D UX-Net: A Large Kernel Volumetric ConvNet Modernizing Hierarchical Transformer for Medical Image Segmentation [Internet]. arXiv; 2022 [cited 2025 Nov 24]. Available from: https://arxiv.org/abs/2209.15076.

18. Wang X, Wang X, Yan Z, Yin F, Li Y, Liu X, Liu Y. Enhanced choroid plexus segmentation with 3D UX-Net and its association with disease progression in multiple sclerosis. Multiple Sclerosis and Related Disorders. 2024;88:105750. doi: 10.1016/j.msard.2024.105750.

19. Eisma JJ, McKnight CD, Hett K, Elenberger J, Han CJ, Song AK, Considine C, Claassen DO, Donahue MJ. Deep learning segmentation of the choroid plexus from structural magnetic resonance imaging (MRI): validation and normative ranges across the adult lifespan. Fluids Barriers CNS. 2024;21:21. doi: 10.1186/s12987-024-00525-9.

20. Isensee F, Schell M, Pflueger I, Brugnara G, Bonekamp D, Neuberger U, Wick A, Schlemmer H, Heiland S, Wick W, et al. Automated brain extraction of multisequence MRI using artificial neural networks. Human Brain Mapping. 2019;40:4952–4964. doi: 10.1002/hbm.24750.

21. Tustison NJ, Cook PA, Holbrook AJ, Johnson HJ, Muschelli J, Devenyi GA, Duda JT, Das SR, Cullen NC, Gillen DL, et al. The ANTsX ecosystem for quantitative biological and medical imaging. Sci Rep. 2021;11:9068. doi: 10.1038/s41598-021-87564-6.

22. Cerri S, Puonti O, Meier DS, Wuerfel J, Mühlau M, Siebner HR, Van Leemput K. A contrast-adaptive method for simultaneous whole-brain and lesion segmentation in multiple sclerosis. NeuroImage. 2021;225:117471. doi: 10.1016/j.neuroimage.2020.117471.

23. Jenkinson M, Beckmann CF, Behrens TEJ, Woolrich MW, Smith SM. Review FSL. NeuroImage. 2012; doi: 10.1016/j.neuroimage.2011.09.015.

24. Yushkevich PA, Piven J, Hazlett HC, Smith RG, Ho S, Gee JC, Gerig G. User-guided 3D active contour segmentation of anatomical structures: Significantly improved efficiency

and reliability. NeuroImage. 2006;31:1116–1128. doi: 10.1016/j.neuroimage.2006.01.015.

25. Lin T-Y, Goyal P, Girshick R, He K, Dollár P. Focal Loss for Dense Object Detection. 2017;

26. Loshchilov I, Hutter F. Decoupled weight decay regularization. 7th International Conference on Learning Representations, ICLR 2019. 2019.

27. Taha AA, Hanbury A. Metrics for evaluating 3D medical image segmentation: analysis, selection, and tool. BMC Med Imaging. 2015;15:29. doi: 10.1186/s12880-015-0068-x.

28. Henschel L, Conjeti S, Estrada S, Diers K, Fischl B, Reuter M. FastSurfer - A fast and accurate deep learning based neuroimaging pipeline. NeuroImage. 2020;219:117012. doi: 10.1016/j.neuroimage.2020.117012.

29. Palaniappan D, Jain R, Premavathi T, Parmar K, Ghribi W, Ahmed AM, Ahmad N. YOLO in Healthcare: A Comprehensive Review of Detection Architectures, Domain Applications, and Future Innovations. IEEE Access. 2025;13:145714–145735. doi: 10.1109/ACCESS.2025.3599358.

## Figures

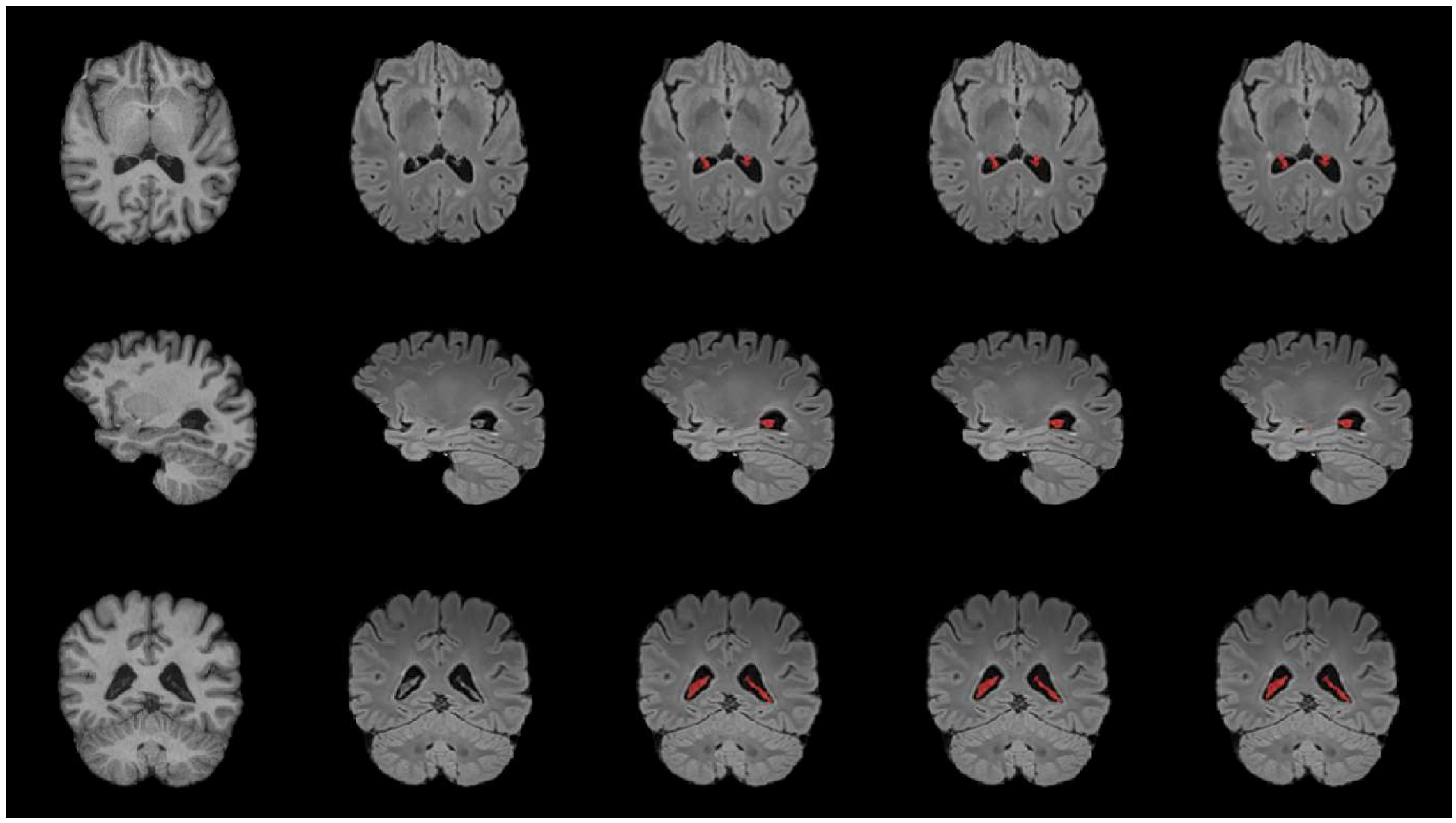

Figure 1. Lateral ventricle choroid plexus segmentation of a person with multiple sclerosis. The images from the left to the right are MPRAGE, FLAIR, the manually corrected mask in red, the SwinUNETR mask and the UXNET mask.

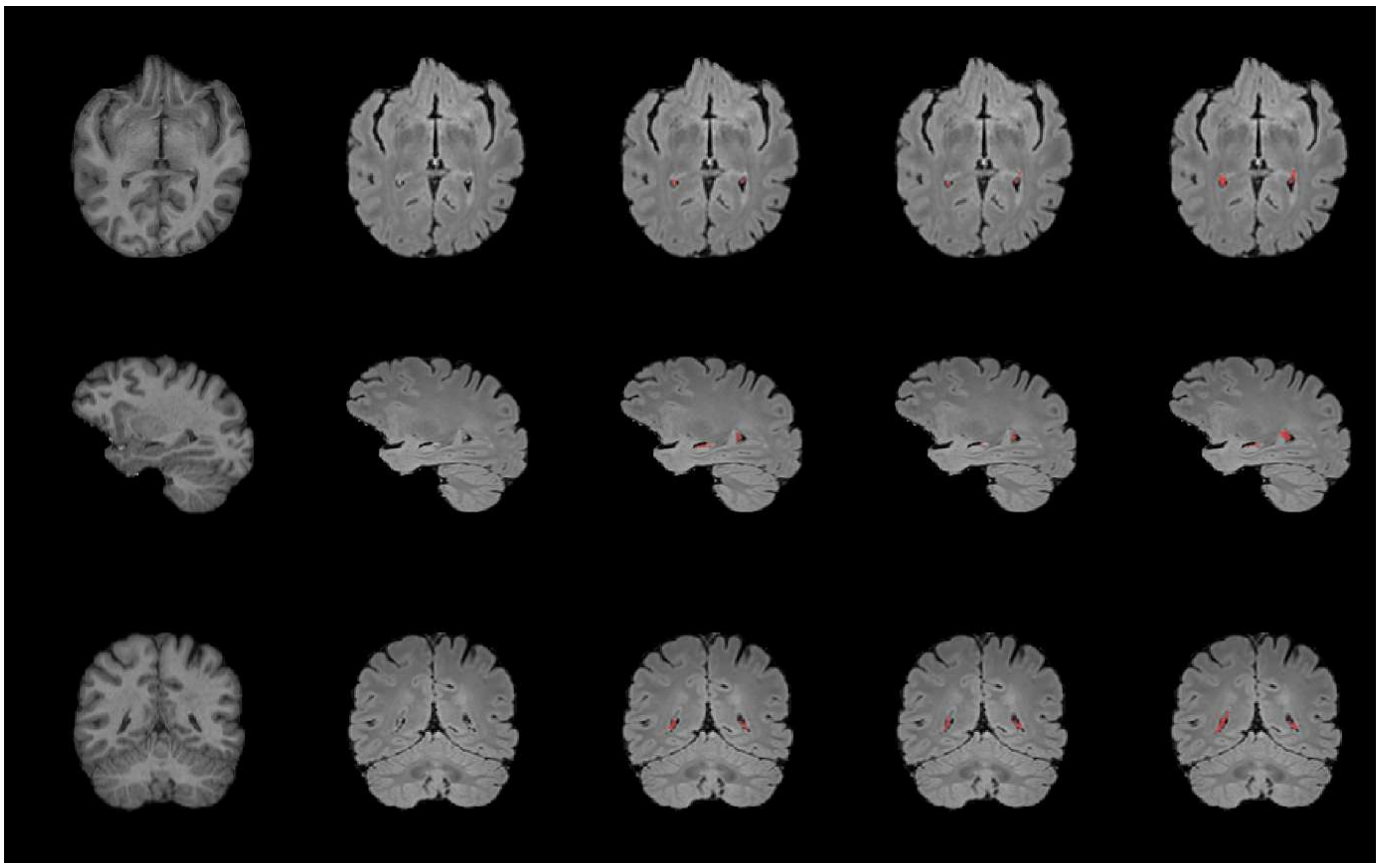

Figure 2. Lateral ventricle choroid plexus segmentation of a person with multiple sclerosis with small lateral ventricles. The images from the left to the right are MPRAGE, FLAIR, the manually corrected mask in red, the SwinUNETR mask and the UXNET mask.